\RequirePackage{amsthm}

\documentclass[pdflatex, sn-basic]{sn-jnl}% Basic Springer Nature Reference Style/Chemistry Reference Style
\usepackage{graphicx}%
\usepackage{multirow}%
\usepackage{amsmath,amssymb,amsfonts}%
\usepackage{amsthm}%
\usepackage{mathrsfs}%
\usepackage[title]{appendix}%
\usepackage{xcolor}%
\usepackage{textcomp}%
\usepackage{manyfoot}%
\usepackage{booktabs}%
\usepackage{algorithm}%
\usepackage{algorithmic}%
\usepackage{listings}%
\usepackage{bm}
\usepackage{soul}
\usepackage{comment}
\usepackage{booktabs}       % professional-quality tables
\usepackage{multirow}
\usepackage{mathtools}

\theoremstyle{thmstyleone}%
\newtheorem{theorem}{Theorem}%  meant for continuous numbers
\newtheorem{proposition}[theorem]{Proposition}% 

\theoremstyle{thmstyletwo}%

\theoremstyle{thmstylethree}%
\newtheorem{definition}{Definition}%

\begin{document}

\title[KDM for Probabilistic Deep Learning]{Kernel Density Matrices for Probabilistic Deep Learning}


\author*[1]{\fnm{Fabio A.} \sur{González}}\email{fagonzalezo@unal.edu.co}
\author[2]{\fnm{Raúl} \sur{Ramos-Pollán}}\email{raul.ramos@udea.edu.co}
\author[1]{\fnm{Alejandro} \sur{Gallego}}\email{jagallegom@unal.edu.co}

\affil*[1]{\orgdiv{MindLab,  Depto. de Ing. de Sistemas e Industrial}, \orgname{Universidad Nacional de Colombia}, \orgaddress{ \city{Bogotá},  \state{DC}, \country{Colombia}}}
\affil[2]{\orgdiv{Depto. Ing. de Sistemas}, \orgname{Universidad de Antioquia}, \orgaddress{ \city{Medellín},  \state{Antioquia}, \country{Colombia}}}


\abstract{This paper introduces a novel approach to probabilistic deep learning, kernel density matrices, which provide a simpler yet effective mechanism for representing joint probability distributions of both continuous and discrete random variables. In quantum mechanics, a density matrix is the most general way to describe the state of a quantum system. This work extends the concept of density matrices by allowing them to be defined in a reproducing kernel Hilbert space. This abstraction allows the construction of differentiable models for density estimation, inference, and sampling, and enables their integration into end-to-end deep neural models. In doing so, we provide a versatile representation of marginal and joint probability distributions that allows us to develop a differentiable, compositional, and reversible inference procedure that covers a wide range of machine learning tasks, including density estimation, discriminative learning, and generative modeling. The broad applicability of the framework is illustrated by two examples: an image classification model that can be naturally transformed into a conditional generative model, and a model for learning with label proportions that demonstrates the framework's ability to deal with uncertainty in the training samples. The framework is implemented as a library and is available at: \url{https://github.com/fagonzalezo/kdm}.}




\maketitle

\section{Introduction}

The density matrix is a powerful mathematical tool used in quantum mechanics to describe the state of a quantum system \citep{Zwiebach2022MasteringApplications}. Unlike the wave function, which provides a complete description of a quantum system in a pure state, the density matrix can describe both pure and mixed states. A mixed state occurs when we have a statistical ensemble (classical uncertainty) of different possible quantum states (quantum uncertainty), rather than a single known state. A density matrix can be thought of as a probability distribution that encodes both the classical and quantum uncertainty of a quantum system. An attractive feature of the density matrix formalism is that it provides a convenient and powerful computational framework that combines probability and linear algebra. For example, by taking the trace of the product of the density matrix with an operator, one can determine the expectation value of that operator. 

Formally, a density matrix is a Hermitian operator that is positive semidefinite, and it acts on a Hilbert space that represents the state space of a quantum system \citep{Nielsen2012QuantumInformation}. This work introduces an abstraction, the kernel density matrix (KDM), which extends the notion of density matrix to reproducing kernel Hilbert spaces (RKHS). This abstraction has several interesting features from a machine learning perspective: first, it can represent both discrete and continuous probability distributions; second, it provides efficient kernelized operations for density computation and inference; third, it can be used to develop models for density estimation, inference, and sampling that are differentiable. Accordingly, it could be integrated into end-to-end deep neural models.

The main contributions of this paper are: (i) the extension of the concept of density matrix to RKHSs in the form of KDMs; (ii) demostrating that KDM is a versatile mechanism for representing both continuous and discrete probability distributions; (iii) developing a method for conducting inference using KDMs that is differentiable, compositional, and reversible; (iv) presenting parametric and non-parametric learning algorithms for shallow and deep models involving KDMs; and (v) exploring the framework's capabilities in diverse learning tasks.

\subsection{Related work}

The model presented in this work can be regarded as a quantum-inspired machine-learning method \citep{Schuld2018SupervisedComputers}. The majority of quantum machine-learning methods rely on wave functions that depict pure states \citep{Cerezo2023ChallengesLearning}. Density matrices provide an attractive way to represent probability distributions in machine learning, but there have been few studies exploring their potential applications in specific machine learning tasks:  classification \citep{Tiwari2019TowardsClassifier, Sergioli2018AClassifier, Giuntini2023Quantum-inspiredClassification}, clustering  \citep{Wolf2006LearningRule}, and multimodal learning \citep{Li2021Quantum-inspiredAnalysis} are some examples.    The two studies most closely related to the present one are those by  \citep{Srinivasan2018LearningModels}  and   \citep{Gonzalez2022LearningFeatures} .   \cite{Srinivasan2018LearningModels} use a kernel mean map to embed rank-1 density matrices in a RKHS. The authors apply their approach to formulate a hidden quantum Markov model (HQMM) and present an algorithm that is based on two-stage regression to learn the parameters of the HQMM.  \cite{Gonzalez2022LearningFeatures} combine density matrices with random Fourier features (RFF) \citep{Rahimi2007RandomMachines} to represent probability distributions, perform inference, and integrate them into deep learning models. The current study presents a more general and efficient formulation than that of \citep{Gonzalez2022LearningFeatures}, by implicitly defining density matrices in RKHS induced by a kernel, and introducing kernelized versions of operations over these density matrices. In contrast to  \citep{Srinivasan2018LearningModels}, the framework presented in this study allows the representation of arbitrary rank density matrices and introduces learning algorithms that can be applied to different models including deep end-to-end architectures.

Probabilistic deep learning (PDL) refers to the combination of deep neural networks and probabilistic models. PDL can model uncertainty on the input data, the output predictions and/or the model parameters. Examples of PDL methods include deep Gaussian processes (DGP) \citep{Damianou2013DeepProcesses}, Bayesian neural networks (BNN) \citep{Goan2020BayesianSurvey}, generative adversarial networks (GANs) \citep{Goodfellow2020GenerativeNetworks}, variational autoencoders \citep{Kingma2013Auto-EncodingBayes},  and autoregressive models. The different methods present different approaches to PDL. DGP models the entire training dataset as a stochastic process, while BNN uses a Bayesian approach to model the neural network parameters.  Some methods focus on a particular task, such as GANs, which are designed for generation, while others are more versatile and can support a wide range of tasks. Normalizing flows (NF) \citep{Rezende2015VariationalFlows, Papamakarios2021NormalizingInference} are an example of the latter and can be applied to tasks such as density estimation, sampling, variational inference, clustering, and classification. The framework presented in this paper takes a similar approach but is based on a different foundation that it shares with quantum mechanics. This provides a new way to think about PDL, presenting original characteristics and possibilities that we believe deserve further investigation.

\section{Density matrices and kernel density matrices}
\subsection{Density matrices}
A wave function, denoted as $|\psi\rangle$, is a complex-valued  function that lives in a Hilbert space, $\mathcal{H}$, and provides a complete description of the state of a `pure' quantum system. The probability of finding a system in a particular state is given by the square of the absolute value of the projection of the wave function onto the desired state (also called the probability amplitude). This is known as the Born rule \citep{Nielsen2012QuantumInformation}. 

In addition to the quantum uncertainty represented by the wave function, it is possible for the system to have classical uncertainty. In this case we say that the system is in a `mixed' state, i.e. a statistical mixture of $N$ different states, $|\psi_i\rangle$, each with an associated probability $p_i$, with $\sum_i^N p_i = 1$. This is represented by a density matrix, $\rho$, which is the more general way to represent the state of a quantum system:

\begin{equation}\label{eq:density-matrix}
\rho = \sum_i^N p_i|\psi_i\rangle\langle\psi_i|
\end{equation}

where $\langle\psi_i|$ represents the conjugate transpose of  $|\psi_i\rangle$.  A density matrix  is a Hermitian, positive-semidefinite matrix with unit trace, defined on the Hilbert space associated with the quantum system\footnote{If the Hilbert space is infinite-dimensional, we use the term 'density operators' instead of 'density matrices'.}.  The Born rule can be extended to compute the probability of finding a system with the state represented by $\rho$ in a state $|\psi\rangle$ after a measurement:

\begin{equation}\label{eq:density-matrix-projection}
    p(|\psi\rangle|\rho)= \text{tr}(|\psi\rangle\langle\psi|\rho)=\langle\psi|\rho|\psi\rangle =\sum_i^N p_i|\langle\psi|\psi_i\rangle|^2
\end{equation}

Density matrices are a powerful mechanism to represent quantum probability distributions and efficiently perform different calculations: outcomes of quantum measurements, expected values, properties of composite systems, and system dynamics, among others \citep{Nielsen2012QuantumInformation}. 

The following section introduces the concept of a KDM, which can be thought of as a density matrix defined in the RKHS induced by a kernel. In this work, KDMs are used to efficiently represent joint probability distributions and to perform inference, among other things. The defined operations are differentiable and therefore integrable into deep learning models.  

\subsection{Kernel density matrices}

\begin{definition}[Kernel density matrix] \label{def:kenel-mixture}
A kernel density matrix over a set $\mathbb{X}$ is a triplet $\rho = (\bm{C},\bm{p},k_\theta)$ where $\bm{C}=\{\bm{x}^{(1)},\dots,\bm{x}^{(m)}\}\subseteq \mathbb{X}$, $\bm{p}=(p_1,\dots,p_m) \in \mathbb{R}^m$ and $k_{\theta}:\mathbb{X} \times \mathbb{X}\rightarrow\mathbb{R}$, such that $\forall \bm{x}\in\mathbb{X}, \ k(\bm{x},\bm{x})=1$, $\forall i \ p_i\ge 0$ and $\sum_{i=1}^n p_i = 1$.

The elements of $\bm C$ are the components of the KDM, and the $p_i$  value represents the mixture weight, or probability, of the component $\bm x_i$. If $\phi:\mathbb R^n \rightarrow \mathcal H$ is the mapping  to the RKHS $\mathcal H$ associated to the kernel $k_\theta$,  $\rho$  represents a density matrix defined as in Eq. \ref{eq:density-matrix} with components $|\psi_i\rangle=|\phi(\bm x^{(i)})\rangle$.   The projection function associated to a KDM $\rho$  is defined as:

\begin{equation}\label{eq:f-projection}
    f_{\rho}(\bm{x})=\sum_{\bm{x}^{(i)} \in \bm{C}} p_i k_\theta^2(\bm{x},\bm{x}^{(i)}) 
\end{equation}
\end{definition}

If we replace $k_\theta^2(\bm{x},\bm{x}^{(i)})$ by  $|\langle\phi(\bm x)|\phi(\bm x^{(i)})\rangle|^2$  in  Eq. \ref{eq:f-projection}, we get the Born rule equation Eq. \ref{eq:density-matrix-projection}.  In quantum mechanics it is common to work with complex Hilbert spaces, in this work we use real-valued kernels over a real domain. This implies that the represented density matrices live in a real Hilbert space.  This framework can be easily extended to the complex case, but the real case is sufficient for our goal of representing classical probability distributions.

A central idea in this work is that the projection function in Eq. \ref{eq:f-projection} can be transformed in a probability density function (PDF) by multiplying it by a normalization constant that depends on the kernel of the KDM:
\begin{equation}\label{eq:f_hat-density}
    \hat{f}_{\rho}(\bm{x})=\mathcal{M}_k
    f_{\rho}(\bm{x})
\end{equation}

\paragraph{Discrete kernel density matrices. } A discrete KDM refers to a KDM in which the corresponding kernel is associated with a RKHS of finite dimension. In this work we will use $\mathbb X = \mathbb R ^n$ along with the cosine kernel 
($
    k_\mathrm{cos}(\bm{x}, \bm{y}) = \frac{<\bm{x}, \bm{y}>}{\sqrt{<\bm{x}, \bm{x}><\bm{y}, \bm{y}>}}
$)
to represent discrete probability distributions. The 
normalization constant for this kernel is $\mathcal{M}_{k_\text{cos}} = 1$. It is possible to use any kernel that satisfies $\forall \bm{x}\in\mathbb{X}, \ k(\bm{x},\bm{x})=1$ and has an associated RKHS of finite dimension.

The following propositions shows that a KDM $\rho_{\mathbf{x}}=(\bm{C},\bm{p},k_\mathrm{cos})$ represents a  probability distribution \footnote{We will use the notation $\rho_{\mathbf{x}}$  to indicate that the KDM $\rho$ represents a probability distribution for the random variable $\mathbf{x}$.} for a random variable $\mathbf{x}$ that can take a finite set of values. 

\begin{proposition}{} \label{prop:cat_distribution}
Let $\rho_{\mathbf{x}}=(\bm{C},\bm{p},k_\mathrm{cos})$ be a KDM over  $\mathbb{R}^n$; let $\mathbb{X}=\{\bm{b}^{(1)},\dots,\bm{b}^{(n)} \} \subset \mathbb{R}^n$  be an orthogonal basis of $\mathbb{R}^n$, then $\{f_{\rho}(\bm{b}^{(i)})\}_{i=1,\dots n}$ is a categorical probability distribution for the random variable  $\mathbf{x} \in \mathbb{X}$.
\end{proposition}

In particular, if we use the canonical basis $\mathbb{X}=\{\bm{e}^{(1)},\dots,\bm{e}^{(n)} \}$, the function in Eq. \ref{eq:f_hat-density},    defines a valid categorical PDF over $\mathbb X$:
\begin{equation}
    \hat{f}_{\rho_{\mathbf x}}(\mathbf x)=
    f_{\rho_{\mathbf x}}(\bm x), \text{for } \bm{x} \in \{\bm{e}^{(i)}\}_{i=1\dots n} \label{eq:f_hat-cos}
\end{equation}
As an example, consider the discrete probability distribution $\bm{p} = (0.2, 0.3, 0.5)$. It  can be represented by the KDM $\rho_0 = (\{(1, 0, 0), (0, 1, 0), (0, 0, 1)\},(0.2, 0.3, 0.5), k_\mathrm{cos})$,  also, it can be represented by the KDM $\rho_1 = (\{(\sqrt{0.2}, \sqrt{0.3}, \sqrt{0.5})\},(1), k_\mathrm{cos})$.  

\paragraph{Continuous kernel density matrices. }
A continuous KDM is a KDM over $\mathbb R ^n$ with a radial basis kernel  $k(x,y) = K(x-y)$ that satisfies: 
\begin{equation} \label{eq:kernel_cont_cond}
\begin{split}
    \int K(x) dx & = 1  \\ 
    \int xK(x) dx & = 0  \\
    \int x^2 K(x) dx & > 0  
\end{split}
\end{equation}

There are different kernels that satisfies the conditions in Eq. \ref{eq:kernel_cont_cond}, including the Gaussian, Epanechnikov and tricube kernels \citep{Wasserman2006AllStatistics}. In this work we will use  the Gaussian (or RBF) kernel, 
$
    k_{\mathrm{rbf}, \sigma}(\bm{x},\bm{y}) = e^{\frac{-||\bm{x} - \bm{y}||^2}{2\sigma^2}}$.
    
A KDM $\rho_\mathbf{x}=(\bm{C},\bm{p},k_\mathrm{rbf})$ with $\bm C \subseteq \mathbb{R}^n$ represents a continuous probability distribution of a random variable $\mathbf{x} \in \mathbb{R}^n$ with a PDF given by Eq. \ref{eq:f_hat-density}  with normalizing constant  $\mathcal{M}_{k_\text{rbf}}=\frac{1}{\sqrt{(2\pi)^n}(\sqrt{2}\sigma)^n}$. It is not difficult to check that the PDF corresponding to $\rho_\mathbf{x}$,  $\hat{f}\rho_{\mathbf x}$,  is a valid PDF over $\mathbb{R}^n$. In fact, Eq. \ref{eq:f_hat-density}, with normalizing constant $\mathcal{M}_{k_\text{rbf}}$, is a generalization of the non-parametric estimation approach to density estimation with kernels known as kernel density estimation (KDE) \citep{Rosenblatt1956RemarksFunction, Parzen1962OnMode}.  As shown in the present work, KDMs can be used in wider range of applications and can be learned using non-parametric and parametric approaches.

\subsection{Density estimation with kernel density matrices}\label{subsect:density-estimation}

Since a KDM extends the notion of kernel density estimator, it is possible to perform non-parametric density estimation with KDMs. In fact, Equation  \ref{eq:f_hat-density} corresponds to a KDE density estimator with a Gaussian kernel and a bandwidth parameter $\sqrt{2}\sigma$. The following theorem  guarantees that the estimator in  Eq. \ref{eq:f_hat-density} converges in probability to the real PDF:
\begin{theorem}[\citet{Parzen1962OnMode}]
Let  $\bm{D}=\{\bm{x}^{(1)}, \dots,\bm{x}^{(\ell)}\}\subseteq \mathbb{R}^n$ be a set of iid samples drawn from a probability distribution with PDF $f$; let   $\rho^\ell_\mathbf{x}=(\bm{D},\bm{p}=\{\frac{1}{\ell},\dots,\frac{1}{\ell}\},k_{\mathrm{rbf}, \sigma_\ell})$  be a KDM; and let $\hat{f}_{\rho^\ell_\mathbf{x}}$ be defined as in  Eq. \ref{eq:f_hat-density}. Assume that $f$ is continuous at $\bm{x}$ and that $\sigma_\ell \rightarrow 0$ and $\ell\sigma_\ell \rightarrow \infty$ as $\ell \rightarrow \infty$. Then $\hat{f}_{\rho^\ell_\mathbf{x}}(x) \xrightarrow{p} f(x)$.
\end{theorem}
The following proposition corresponds to the analogous result for categorical probability distributions.
\begin{proposition}\label{prop:convergence-discrete-kqm}
Let  $\mathbb{X}=\{\bm{e}^{(1)},\dots,\bm{e}^{(n)} \}$  be the canonical basis of  $\mathbb R^n$; let  $\bm{D}=\{x_1, \dots,x_\ell\}\subseteq \mathbb R^n $  be a set of iid samples drawn from $\mathbb X$ with categorical probability distribution $\bm{p}=(p_1,\dots,p_n)$; let   $\rho^\ell_\mathbf{x}=(\bm{D},\bm{p}=(\frac{1}{\ell},\dots,\frac{1}{\ell}),k_{\mathrm{cos}})$  be a KDM; and let $\hat{f}_{\rho^\ell_\mathbf{x}}(i)$ be defined as in  Eq. \ref{eq:f_hat-cos}.  Then $\hat{f}_{\rho_\ell}(i) \rightarrow p_i$ as $\ell \rightarrow \infty$.
\end{proposition}

It is also possible to perform parametric density estimation by maximizing the likelihood of the KDM parameters given a training dataset. Given a KDM with $K$ components $\rho=(\bm{C}=\{\bm{c}^{(1)},\dots,\bm{c}^{(K)}\},\bm{p}=\{p_1,\dots,p_K\},k_{\theta})$, where $\theta$ indicate a set of parameters of the kernel, and a training data set $\bm{D}=\{\bm{x}_1, \dots, \bm{x}_\ell\}$, solving the following optimization problem:
\begin{equation}\label{eq:density-opt-problem}
    \max_{\bm{C}, \bm{p},\theta} \sum_{i=1}^\ell \log \hat{f}_\rho(x_i)
\end{equation}
will find a KDM that maximizes the probability density of the samples in $\bm{D}$. For the cosine or RBF kernels this is equivalent to perform maximum likelihood estimation.\footnote{The optimization problem in Eq. \ref{eq:density-opt-problem} could be ill-posed, e.g. for the RBF kernel,  Eq. \ref{eq:density-opt-problem} could get arbitrary large values if one sample $\bm x_i$  coincides with one of the KDM components and the $\sigma$ parameter of the kernel goes to zero. This could be dealt with using techniques such as restricted optimization or regularization.}

\subsection{Joint densities with kernel density matrices}\label{subsect:joint-densities}
In quantum mechanics, the state of a bipartite system with subsystems $A$ and $B$  is represented by a density matrix in the tensor product Hilbert space $\mathcal H_A\otimes \mathcal H_B$, where $\mathcal H_A$ and $\mathcal H_B$ are the representation spaces of $A$ and $B$ respectively \citep{Nielsen2012QuantumInformation}. Following the same line of thought, a joint KDM for random variables  $\mathbf{x} \in \mathbb{X}$ and $\mathbf{y} \in \mathbb{Y}$
is defined as follows: $\rho_{\mathbf{x}, \mathbf{y}} = (\bm{C},\bm{p},k_{\mathbb{X}} \otimes k_{\mathbb{Y}})$ where $\bm{C}\subseteq \mathbb{X}\times\mathbb{Y}$ ,  $k_{\mathbb{X}}$ and $k_{\mathbb{Y}}$ are kernels over  $\mathbb{X}$ and $\mathbb{Y}$ respectively, and  $k_\mathbb{X}\otimes k_\mathbb{Y}((\bm{x},\bm{y}),(\bm{x}',\bm{y}'))=k_\mathbb{X}(\bm{x},\bm{x}')k_\mathbb{Y}(\bm{y},\bm{y}')$. If $\mathcal{M}_{k_{\mathbb X}}$  and $\mathcal{M}_{k_{\mathbb Y}}$ are the normalizing constants corresponding to the kernels  $k_\mathbb{X}$ and  $k_\mathbb{Y}$  respectively, the normalizing constant for $k_\mathbb{X}\otimes k_\mathbb{Y}$ in Eq. \ref{eq:f_hat-density} is $\mathcal{M}_{k_\mathbb{X}\otimes k_\mathbb{Y}}=\mathcal{M}_{k_{\mathbb X}}\mathcal{M}_{k_{\mathbb Y}}$.
It is also possible to easily calculate a marginal KDM from a joint KDM. Suppose that $k_\mathbb{X}=k_\mathrm{rbf}$ , then a marginal KDM $\rho_{\mathbf{y}}$ must satisfy:

\begin{equation}
    \begin{split}
        \hat{f}_{\rho_{\mathbf{y}}}(\bm y) & = \int \hat{f}_{\rho_{\mathbf{x}, \mathbf{y}}}(\bm x, \bm y) d\bm x  \\ 
          & = \int \sum_i p_i \mathcal{M}_{k_{\mathbb X}} \mathcal{M}_{k_{\mathbb Y}} k^2_\mathbb{X}(\bm x^{(i)}, \bm x)k^2_\mathbb{Y}(\bm y^{(i)}, \bm y)d \bm x  \\ 
          & = \sum_i p_i \mathcal{M}_{k_{\mathbb Y}} k^2_\mathbb{Y}(\bm y^{(i)}, \bm y) 
    \end{split}     \label{eq:marginal-kernel-mixture}      
\end{equation}

It is not difficult to see that $\rho_\mathbf{y} = (\{\bm y^{(i)} | (\bm x^{(i)}, \bm y^{(i)}) \in \bm{C}\},\bm{p},k_\mathbb{Y})$satisfies \ref{eq:marginal-kernel-mixture}. The same argument applies for defining  $\rho_{\mathbf{x}}$. For $k_\mathbb{X}=k_\mathrm{cos}$, we replace the integral in Eq. \ref{eq:marginal-kernel-mixture} by a sum over a basis (see Proposition \ref{prop:cat_distribution}) to obtain an analogous result.

\subsection{Inference with kernel density matrices} \label{subsect:inference}
Inference aims to estimate unknown output variables from known input variables and a model's parameters. A probabilistic approach models the input-output relationship as a probability distribution, e.g., $p(\mathbf x', \mathbf y')$, which reflects uncertainty about the training data-generation process. Additional uncertainty may arise from imprecise input data during prediction, which can be modeled as a probability distribution over the input domain, $p(\mathbf{x})$. Note that we use a different random variable, $\mathbf x$, to represent a new input fed to the trained model during prediction, than the random variable used to represent input training samples, $\mathbf x'$ . In general, $ p(\mathbf x) \neq \int p(\mathbf x', \mathbf y'=\bm y)d\bm y$. When predicting output variables, both sources of uncertainty need to be considered and reflected in the output distribution, $p(\mathbf y)$. Inference transforms a probability distribution of input variables, $p(\mathbf x)$, into a distribution of output variables, $p(\mathbf y)$, using a joint probability of inputs and outputs, $p(\mathbf x', \mathbf y')$.

We will demonstrate how this process can be modeled using KDMs. Each one of the probability distributions $p(\mathbf{x})$, $p(\mathbf{x}', \mathbf{y}')$ and $p(\mathbf{y})$ are represented by KDMs as follows:
\begin{align}
    \rho_{\mathbf{x}} &=(\{\bm{x}^{(i)}\}_{i=1\dots m},(p_i)_{i=1\dots m}, k_\mathbb{X}) \label{eq:rho_x}\\
    \rho_{\mathbf{x',y'}}  &=(\{(\bm{x}'^{(i)},\bm{y}'^{(i)})\}_{i=1\dots m'},(p'_i)_{i=1\dots m'}, k_\mathbb{X} \otimes k_\mathbb{Y}) \label{eq:rho_xy}\\
    \rho_{\mathbf{y}} &=(\{\bm{y}'^{(i)}\}_{i=1\dots m'},(p''_i)_{i=1\dots m'}, k_\mathbb{Y}) \label{eq:rho_y}
\end{align}

Notice that if we have a unique input sample $\bm{x}$,   $\rho_{\mathbf{x}}$ would have only one component. In general,  we have a probability distribution represented by the KDM $\rho_{\mathbf{x}}$ with $m$ components. Also, note that $k_\mathbb{X}$ in $\rho_{\mathbf{x}}$, is the same kernel as in  $\rho_{\mathbf{x}',\mathbf{y}'}$. Likewise,  $k_\mathbb{Y}$ in $\rho_{\mathbf{y}}$, is the same kernel as in  $\rho_{\mathbf{x}',\mathbf{y}'}$. Also, the $\bm{y}'^{(i)}$ components of the KDM $\rho_{\mathbf{y}}$ are same as the ones of the KDM $\rho_{\mathbf{x}',\mathbf{y}'}$. The probabilities of the inferred  $\rho_{\mathbf{y}}$ KDM are given by the following expression:

\begin{equation} 
\label{eq:inference-probability}
       p''_i =  \sum_{\ell=1}^m\frac{p_\ell p'_i (k_x(\bm{x}^{(\ell)},\bm{x}'^{(i)}))^2}{\sum_{j=1}^{m'}  p'_j (k_x(\bm{x}^{(\ell)},\bm{x}'^{(j)}))^2},   \text{ for } i=1\dots m' 
\end{equation}

\begin{algorithm}[tbh]
\caption{Inference with kernel density matrices}
\label{alg:inference}
    \textbf{input:} Input KDM
    $\rho_{\mathbf{x}} =(\{\bm{x}^{(i)}\}_{i=1\dots m},(p_i)_{i=1\dots m}, k_\mathbb{X})$, 
    joint KDM 
    $\rho_{\mathbf{x',y'}} = (\{(\bm{x}'^{(i)},\bm{y}'^{(i)})\}_{i=1\dots m'},(p'_i)_{i=1\dots m'}, k_\mathbb{X} \otimes k_\mathbb{Y})$
    \begin{algorithmic}[1]
    \STATE  $ \bm{C}_{\mathbf{y}} \gets  \{\bm{y}'^{(i)}\}_{i=1\dots m'}$
    \STATE  $ \bm{p}_{\mathbf{y}} \gets  (p''_k)_{k=1\dots m'}$ where $p''_k$ is calculated using Eq. \ref{eq:inference-probability}
    \STATE {\textbf{return}} $\rho_{\mathbf{y}} = (\bm C_{\mathbf y},\bm p_{\mathbf y},k_{\mathbb Y})$
\end {algorithmic}
\end {algorithm}

The time complexity of the inference process (Algorithm \ref{alg:inference}) based on Eq. \ref{eq:inference-probability} is $O(mm'n)$ where $n$ is the dimension of the input space. This inference procedure is similar to Nadaraya-Watson kernel regression (NWKR) \citep{Nadaraya1964OnRegression, Watson1964SmoothAnalysis}, which uses a kernel function to assign local weights to each output training sample according to the input sample's similarity with each input training sample. However, NWKR only produces a point estimate of the output variable, while we obtain a full probability distribution represented as a KDM. 

In addition to its expressiveness and flexibility,  KDM-based inference has two advantages:
\begin{itemize}
    \item  Compositionality.  Given two joint KDMs $\rho_{\mathbf{x',y'}}$  and $\rho_{\mathbf{y', z'}}$  and input KDM $\rho_{\mathbf{x}}$ , we can infer $\rho_{\mathbf{y}}$ from $(\rho_{\mathbf{x',y'}},\rho_{\mathbf{x}})$  and then infer $\rho_{\mathbf{z}}$ from  $(\rho_{\mathbf{y',z'}},\rho_{\mathbf{y}})$.
    \item Reversibility. Thanks to the symmetry  of the joint KDMs $\rho_{\mathbf{x',y'}}$, it can be used to infer  $\rho_{\mathbf{x}}$ from  $\rho_{\mathbf{y}}$  as well as to infer  $\rho_{\mathbf{y}}$ from  $\rho_{\mathbf{x}}$ .
\end{itemize}

From the point of view of quantum mechanics,  $\rho_{\mathbf{x',y'}}$ is a density matrix representing the state of a bipartite system, and the inference process corresponds to a measurement operation that collapses the $\mathbf{x}$ subsystem to a state $\rho_{\mathbf x}$. This collapse affects the $\mathbf y$ subsystem whose state is changed to be $\rho_\mathbf{y}$. With density matrices, this process corresponds to applying a collapse operator to  $\rho_{\mathbf{x',y'}}$ and computing a partial trace \citep{Gonzalez2022LearningFeatures}. With our KDM representation, this process is efficiently done by Eq. \ref{eq:inference-probability}.

The following proposition shows that the probabilities assigned by Eq. \ref{eq:inference-probability} generate a KDM that is equivalent to the one calculated by Bayesian inference. 

\begin{proposition}\label{prop:kqm-inference}
    Let $\mathbf{x}'$ and $\mathbf{y}'$ be random variables with  a joint probability distribution represented by a KDM $\rho_{\mathbf{x',y'}}$ (Eq. \ref{eq:rho_xy}), and let  $\rho_{\mathbf{x}}$ be the KDM defined by  Eq. \ref{eq:rho_x}), then the KDM $\rho_{\mathbf{y}}$ (Eq. \ref{eq:rho_y} and Eq. \ref{eq:inference-probability}) represents a predicted probability distribution with PDF:
    \begin{equation}
    \hat{f}_{\rho_{\mathbf{y}}}(\bm{y}) = \sum_{\ell=1}^m p_\ell \hat{f}_{\rho_{\mathbf{x',y'}}}(\bm y|\bm x^{(\ell)}) 
\end{equation}
    where 
    \begin{itemize}
        \item $\hat{f}_{\rho_{\mathbf{x',y'}}}(\bm y|\bm x^{(\ell)})=\frac{\hat{f}_{\rho_{\mathbf{x',y'}}}(\bm x^{(\ell)}, \bm y)}{\int \hat{f}_{\rho_{\mathbf{x',y'}}}(\bm x^{(\ell)}, \bm y)d\bm y}$   for  $k_\mathbb Y = k_\text{rbf}$
\item $\hat{f}_{\rho_{\mathbf{x',y'}}}(\bm y|\bm x^{(\ell)})=\frac{\hat{f}_{\rho_{\mathbf{x',y'}}}(\bm x^{(\ell)}, \bm y)}{\sum_{i=1}^n \hat{f}_{\rho_{\mathbf{x',y'}}}(\bm x^{(\ell)}, \bm e^{(i)})}$   for  $k_\mathbb Y = k_\text{cos}$
    \end{itemize}

\end{proposition}

The parameters of the inference model correspond to the parameters of the KDM  $\rho_{\mathbf{x',y'}}$. This parameters can be estimated in three different ways:
\begin{itemize}
    \item Non-parametric. Use the training dataset as the parameters of  $\rho_{\mathbf{x',y'}}$  as described in Sect. \ref{subsect:density-estimation}. This leads to a memory-based learning method that may not scale well to large training datasets.
    \item Maximum likelihood learning. Estimate the parameters of $\rho_{\mathbf{x',y'}}$  by maximizing  the probability density assigned by $\rho_{\mathbf{x',y'}}$  to the training dataset (Equation \ref{eq:density-opt-problem}) (Algorithm \ref{alg:MLE-training}). 
    \item Discriminative learning. Use Eq. \ref{eq:inference-probability} to perform a forward pass of an input KDM $\rho_{\mathbf{}{x}}$ that can represent an individual sample (no uncertainty) or a distribution over the input space (uncertainty). Then, we minimize a suitable loss function (such as cross-entropy for classification or mean squared error for regression) with respect to the model parameters using gradient descent (Algorithm \ref{alg:discriminative-training}).
\end{itemize}

For discriminative learning, the loss function depends on the kind of output variable, which is determined by the kernel $k_{\mathbb{Y}}$.  Given  an output KDM  $\rho_{\mathbf y} = (\{\bm{y}'^{(i)}\}_{i=1\dots m'},(p''_i)_{i=1\dots m'}, k_\mathbb{Y})$, where $k_{\mathbb{Y}}=k_{\mathrm{cos}}$, and a real output label  $\bm y \in \mathbb{R}^n$, with $\sum_{j=1}^n y_j=1$ we can use, for instance, a cross-entropy loss:
\begin{equation}\label{eq:xent-loss}
    \mathcal{L}_{\text{xe}}(\rho_{\bm y},\bm y) = \textit{CrossEntropy}(\bm \pi, \bm y),
\end{equation}
where 
\begin{equation}\label{eq:categorical-from-mixture}
\bm \pi_j = \sum_{i=1}^{m'} p''_i(\bm{y}_j^{(i)})^2  \text{, for } 
j=1\dots n
\end{equation}
When $k_{\mathbb{Y}}=k_{\mathrm{rbf}}$, we have a regression problem, in this case we can use a loss function such as mean square error:
\begin{equation}
    \mathcal{L}_{\text{mse}}(\rho_{\bm y},\bm y^{(i)}) = \textit{mse}(\hat{\bm y}, \bm y^{(i)}),
\end{equation}
where $\hat{\bm y} = \sum_{j=1}^{m'} p''_i \bm y'^{(i)}_j $ .
%%%%%%%%%%%%%%%%%%%%%%%%%%%%%%%%%%%%%%%%%%%%%%%%%%%%%%%%%%%%

\begin{algorithm}[tbh]
\caption{Maximum likelihood training}
\label{alg:MLE-training}
 \textbf{input:}   training dataset $\bm D = \{(\bm{x}^{(i)},\bm{y}^{(i)})\}_{i=1\dots \ell}$,
    kernels  $k_{\mathbb X}$ and $k_{\mathbb Y}$,
    number of components $m'$ 
\begin{algorithmic}[1]
    \STATE Initialize $\bm{C_{\mathbf{x'y'}}} = \{(\bm{x'}^{(i)},\bm{y'}^{(i)})\}_{i=1\dots m'}$ with a random sample from $\bm D$
    \STATE Initialize $\bm{p}_{\mathbf {x'y'}} \gets (\frac{1}{m'})_{i=1\dots m'}$
    \STATE $\rho_{\mathbf{x',y'}} = (\bm C_{\mathbf {x'y'}},\bm p_{\mathbf {x'y'}},k_{\mathbb X}\otimes k_{\mathbb Y})$
    \STATE Use gradient-based optimization to find:
    \begin{equation*}
                \max_{\bm C_{\mathbf {x'y'}}, \bm p_{\mathbf {x'y'}} ,\theta} \sum_{i=1}^\ell \log \hat{f}_{\rho_{\mathbf {x'y'}}}(\bm{x}^{(i)}, \bm{y}^{(i)}) 
    \end{equation*}
    \STATE \textbf{return} $\rho_{\mathbf{x',y'}} = (\bm C_{\mathbf {x'y'}},\bm p_{\mathbf {x'y'}},k_{\mathbb X}\otimes k_{\mathbb Y})$
\end {algorithmic}
\end {algorithm} 

\begin{algorithm}[tbh] 
\caption{Discriminative training}
\label{alg:discriminative-training}
\textbf{input:} training dataset $\bm D = \{(\bm{x}^{(i)},\bm{y}^{(i)})\}_{i=1\dots \ell}$,
    kernels  $k_{\mathbb X, \sigma}$ and $k_{\mathbb Y}$
    number of components $m'$, 
    loss function $\mathcal{L}$
\begin{algorithmic}[1]
    \STATE Initialize $\bm{C_{\mathbf {x'y'}}}  \gets  \{(\bm{x'}^{(i)},\bm{y'}^{(i)})\}_{i=1\dots m'}$ with a random sample from $\bm D$
    \STATE \quad $\bm{p}_{\mathbf {x'y'}} \gets (\frac{1}{m'})_{i=1\dots m'}$
    \STATE \quad $\rho_{\mathbf{x',y'}}  \gets  (\bm C_{\mathbf {x'y'}},\bm p_{\mathbf {x'y'}},k_{\mathbb X, \sigma}\otimes k_{\mathbb Y})$
    \STATE \quad $\rho^{(i)}_\mathbf{x} \gets (\{\bm{x}^{(i)}\}, (1), k_{\mathbb{X},\sigma})$ for all $i=1\dots \ell$ \COMMENT{create  a KDM for each input sample}
    \REPEAT 
    \STATE Calculate $\rho^{(i)}_\mathbf{y}$ using  Eq. \ref{eq:inference-probability} with parameters $\rho^{(i)}(\mathbf{x})$ and  $\rho({\mathbf{x',y'}})$ for all $i=1\dots \ell$
    \STATE Perform a step of gradient-based method to optimize:
    $                \min_{\bm C_{\mathbf{x',y'}} , \bm p_{\mathbf{x',y'}} ,\sigma} \sum_{i=1}^n  \mathcal{L}(\rho_{\mathbf y}^{(i)}, \bm{y}^{(i)})$
    
    \UNTIL stop criteria is met
    \STATE \textbf{return} $\rho_{\mathbf{x',y'}} = (\bm C_{\mathbf {x'y'}},\bm p_{\mathbf {x'y'}},k_{\mathbb X}\otimes k_{\mathbb Y})$
\end {algorithmic}
\end {algorithm}

\subsection{Sampling from kernel density matrices}\label{subsect:sampling}
Sampling from a KDM $\rho_{\mathbf{x}} =(\{\bm{x}^{(i)}\}_{i=1\dots m},  \allowbreak (p_i)_{i=1\dots m},  \allowbreak  k_{\mathrm{rbf}, \sigma})$ with $\bm{x}^{(i)} \in \mathbb{R}^n$ can be performed in two steps. First, draw a sample $i'$ from a categorical distribution with probabilities $(p_i)_{i=1\dots m}$. Second, draw a sample from a normal distribution $\mathcal N(\bm{x}^{(i')}, 2\sigma\bm I_n)$, where $\bm{I_n}$ is the $n$-dimensional identity matrix.

To sample from a KDM $\rho_{\mathbf{x}} =(\{\bm{x}^{(i)}\}_{i=1\dots m},  \allowbreak (p_i)_{i=1\dots m},  \allowbreak  k_{\mathrm{cos}})$, we draw a sample from a categorical distribution with probabilities given by $(\bm{\pi_j})_{j=1\dots n}$, where $\bm{\pi_j}$ is defined in the same way as in Eq. \ref{eq:categorical-from-mixture}.

Sampling layers that sample from a KDM can also be included in a deep model. To maintain differentiability, we can use the Gumbel-softmax reparameterization for sampling categorical variables, as proposed by \citep{Jang2017CategoricalGumbel-softmax}, along with the well-known reparameterization trick for Gaussian distributions. Another alternative is to use implicit reparameterization gradients, \citep{Figurnov2018ImplicitGradients}.

\section{Experiments}

\begin{figure*}[t]
    \centering
    \includegraphics[width=0.98\textwidth]{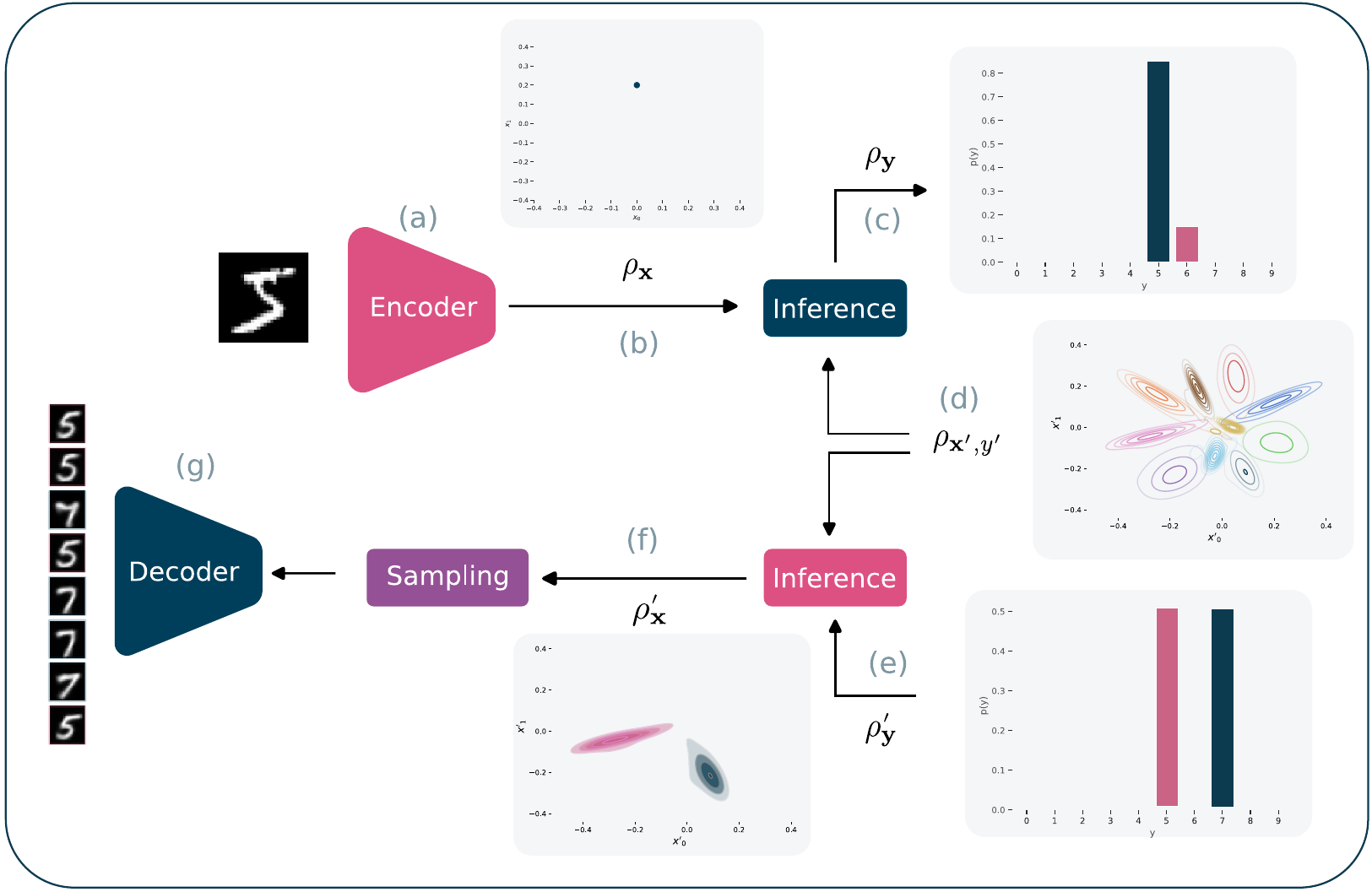}
    \caption{classification and generation with KDMs. The top part represents a predictive model that uses an encoder (a) to map input samples into a latent space; the output of the encoder is represented as a KDM $\rho_{\mathbf x}$ (b)  with one component, which is used to infer an output probability distribution of labels, represented by a KDM  $\rho_{\mathbf y}$ (c),   using Eq. \ref{eq:inference-probability} ; the classifier model has as a parameter a joint distribution of inputs and outputs, represented by a KDM $\rho_{\mathbf x', \mathbf y'}$ (d), which is learned with algorithm \ref{alg:discriminative-training}.  The joint probability can be used to do conditional generation, as depicted in the bottom of the diagram. In this case, the input is a distribution of labels represented by the KDM  $\rho'_{\mathbf y}$ (e) , which along with   $\rho_{\mathbf x', \mathbf y'}$  (d) is used to infer a predicted KDM  $\rho'_{\mathbf x}$  using Eq. \ref{eq:inference-probability}; we sample (f) from this KDM to generate input samples in the latent space which are decoded (g) to the original input space.}
    \label{fig:classification_generation}
\end{figure*}

The purpose of this section is to demonstrate the versatility and flexibility of the framework with two different learning tasks. Rather than focusing on improving the performance of state-of-the-art methods, these experiments illustrate the special features of the framework and the novel perspective it offers for PDL modeling. First, we present an example of a classification model based on the framework, highlighting the reversibility of KDM-based inference, which allows the model to function as both a classification and a conditional generative model. In the second example, we address the problem of learning from label proportions, where we lack individual training sample labels and have only the proportions of labels in sample bags. We treat this problem as an instance of input training sample uncertainty modeled with KDMs.

\subsection{Bidirectional classification and generation with kernel density matrices}\label{subsec:classification_with_quantum_kernel_mixtures}

Figure \ref{fig:classification_generation} shows the task setup and models explored in this subsection. The upper part shows a classifier that combines a deep encoder, $\Phi_\theta:\mathbb X \rightarrow \mathbb R^n$, and a KDM inference module.  The parameters of the model correspond to the joint KDM $\rho_{\mathbf{x',y'}}  =  (\bm C_{\mathbf {x'y'}},\bm p_{\mathbf {x'y'}},k_{\mathrm{rbf} , \sigma}\otimes k_{\mathrm{cos}})$, which represents the joint probability of the training input  and output samples, and the parameters of the encoder, $\theta$. These parameters, together with the parameters of the encoder, are learned using an extended version of Algorithm \ref{alg:discriminative-training} by minimizing the loss given by Eq. \ref{eq:xent-loss}.

The lower part of Figure \ref{fig:classification_generation}  shows a conditional-generative model that exploits the reversibility of the KDM inference process (Eq. \ref{eq:inference-probability}).  Thanks to the symmetry of $\rho_{\mathbf{x',y'}}$, it is possible to use it to infer a probability distribution of inputs, $\rho_{\mathbf x}$, given a probability distribution of outputs  $\rho_{\mathbf y}$. From this KDM we can draw samples (Subsect. \ref{subsect:sampling}) in the latent space, which can be mapped to the input space using a decoder $\Psi_{\theta'}:\mathbb R^n \rightarrow \mathbb X$. To improve the quality of the generated samples, the joint KDM $\rho_{\mathbf{x',y'}}$ is fine-tuned using maximum likelihood learning (see Subsect. \ref{subsect:inference}) while keeping the parameters $\theta$ of the encoder fixed.  The parameters $\theta'$ of the decoder are independently learned by training a conventional auto-encoder composed of $\Phi_\theta$ and $\Psi_{\theta'}$, with the encoder parameters $\theta$ fixed.

Table \ref{tab:classification-results} shows the  performance of the model on three benchmark datasets: MNIST, Fashion-MNIST and CIFAR-10.  All models use a similar  encoder architecture: a convolutional neural network (3 convolutional layers for MNIST and Fashion-MNIST, and 6 convolutional layers for CIFAR-10). As a baseline, we have coupled the encoder with a dense layer with approximately the same number of parameters as the KDM inference module (Encoder+Dense), so that we can compare KDM with a counterpart model of similar complexity. We also report the classification performance of the model before (KDM) and after performing the maximum likelihood fine tuning (ML-KDM) on the joint KDM $\rho_{\mathbf{x',y'}}$ for improving generation. Hyperparameter tuning of the models was performed using cross-validation on the training partition for each dataset. Ten experiments were run with the best hyperparameters, and  the mean accuracy is reported along with the  $99\%$ confidence interval of a t-test. The results show that the performance of the KDM based model is on par with a deep model of comparable complexity.

\begin{table}[t]
\caption{Classification accuracy comparison of the models on the three datasets: KDM classification model (KDM), KDM model fined-tuned for generation with maximum likelihood learning (ML-KDM), a baseline model with the same encoder coupled with a dense layer. Observe that the classification performance ML-KDM slightly degrades with respect  to KDM, since a more faithful modeling of the input space distribution impacts the discriminative performance. }
\small
  \centering
  \begin{tabular}{llll}
    \toprule
    Dataset     & KDM     & ML-KDM &  Enc.+Dense \\
    \midrule
    Mnist &  $0.993\pm0.001$  & $0.993\pm0.001$ & $0.992\pm0.001$    \\
    FMnist     &  $0.916\pm0.004$    & $0.895\pm0.004$  & $0.907\pm0.005$      \\
    Cifar-10     &    $0.811\pm0.005$    & $0.776\pm0.012$ & $0.810\pm0.006$  \\
    \bottomrule
  \end{tabular}
  \label{tab:classification-results}
\end{table}

\begin{figure} 
    \centering
    \includegraphics[width=0.7\columnwidth]{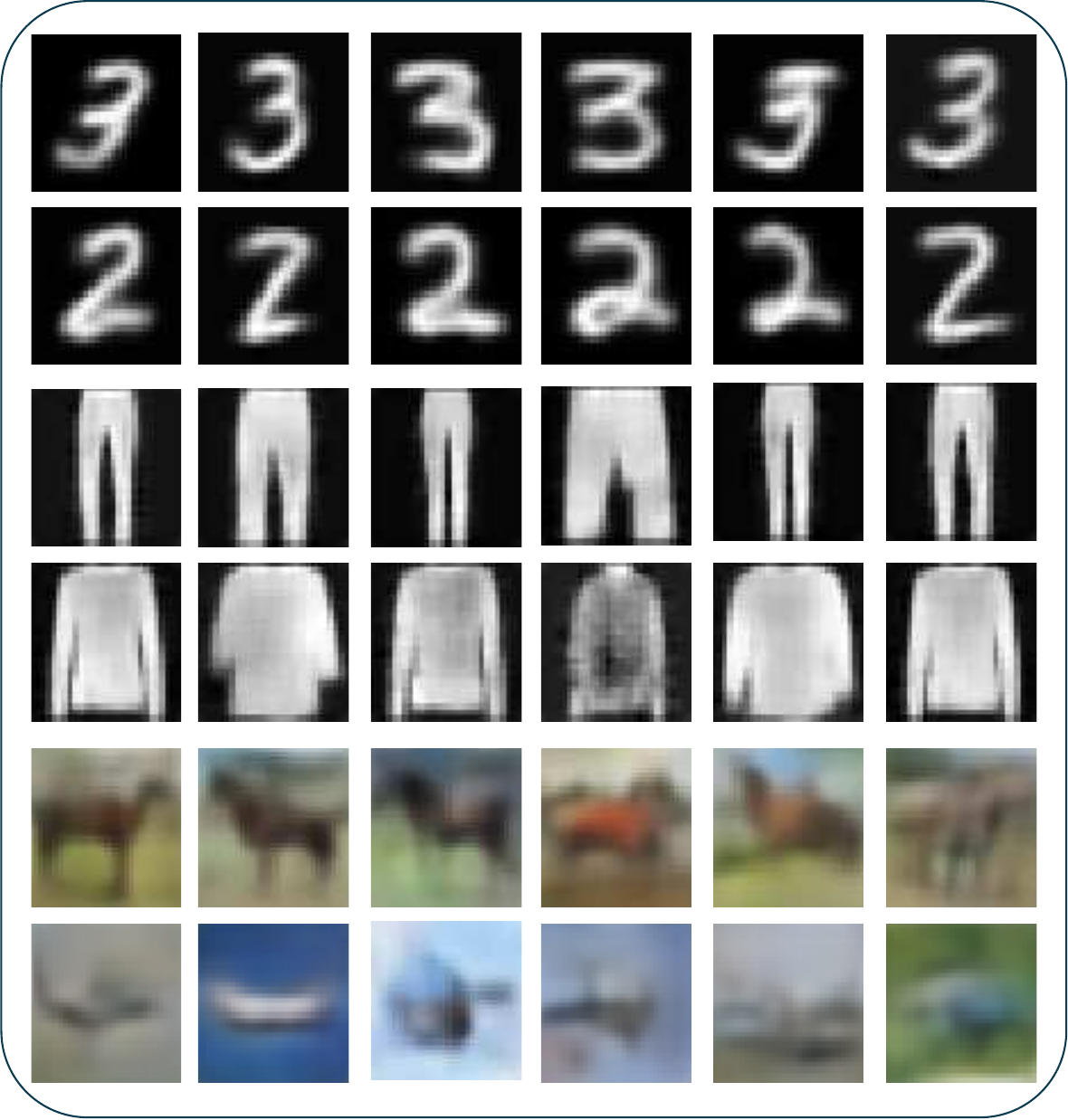}
    \caption{Conditional image generation from Mnist, Fashion-Mnist, and Cifar-10 using the KDM conditional generative model, each row corresponds to a different class.}
    \label{fig:generated-samples}
\end{figure}

The main advantage of the KDM-based model is the probabilistic modeling of the interaction between inputs and outputs, which is exploited by the conditional generative model. It uses the parameters learned by the ML-KDM to conditionally generate samples from the inferred probability distribution of the inputs. Some examples of conditionally generated samples for the three datasets are shown in Figure \ref{fig:generated-samples} .

\subsection{Classification with label proportions}

\begin{figure*}[t]
    \centering
    \includegraphics[width=0.5\textwidth]{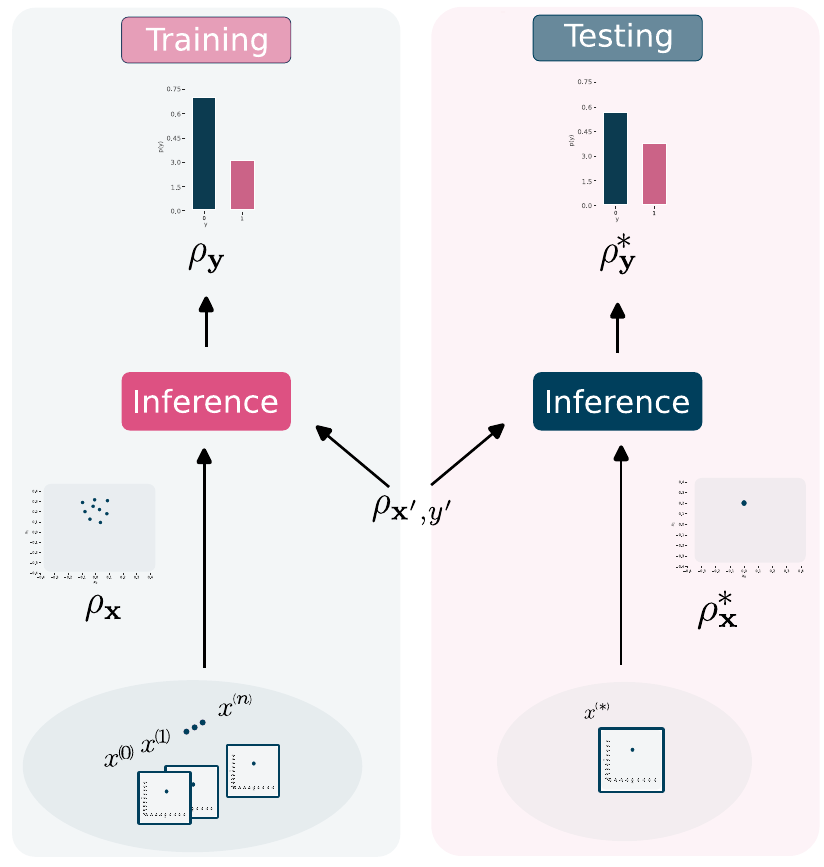}
    \caption{KDM model for classification with label proportions. During training, the model receives as input  bags of instances  $\bm X^{(i)}={(\bm{x}^{(i)j}})_{j=1\dots m_i}$. The training dataset corresponds to a set of pairs $\bm D = {(\bm{X}^{(i)},\bm{y}^{(i)})}_{i=1\dots \ell}$, where each $\bm y^{(i)}$  is a vector representing the label proportions of the $i$-th bag. Each input is represented by a KDM with $m_i$ components. The algorithm learns a joint KDM $\rho_{\mathbf x', \mathbf y'}$. During prediction, the model receives individual samples, $x^{(*)}$, (equivalent to bags with only one element). The algorithm outputs a KDM $\rho_y$.}
    \label{fig:llp}
\end{figure*}

Learning from label proportions (LLP) is a type of weakly supervised problem \citep{Scott2020LearningFramework,Zhang2022LearningNoise}. In this scenario, instead of having explicit class labels for each training instance, we only have class proportions for subsets of training instances, called  bags. The goal of the learner is to build a classifier that accurately labels individual instances. A bag of instances with label proportions can be thought of as a training sample that has uncertainty about the inputs and outputs, as discussed in Subsection \ref{subsect:inference}. This uncertainty can be represented by KDMs as shown in Figure \ref{fig:llp}. During training, the model receives as input  bags of instances  $\bm X^{(i)}={(\bm{x}^{(i)j}})_{j=1\dots m_i}$. The training dataset corresponds to a set of pairs $\bm D = {(\bm{X}^{(i)},\bm{y}^{(i)})}_{i=1\dots \ell}$, where each $\bm y^{(i)}$  is a vector representing the label proportions of the $i$-th bag. Each input bag is represented by a KDM with $m_i$ components.  The algorithm \ref{alg:discriminative-training} can be adapted to handle this bag representation by changing line 4, the resulting algorithm is shown in (Algorithm \ref{alg:discriminative_train_bags}).  The algorithm learns a joint KDM $\rho_{\mathbf x', \mathbf y'}$. During prediction, the model receives individual samples, $x^{(*)}$,  which can be seem as bags with only one element. They are represented by a KDM $\rho^{(*)}_\mathbf{x} = (\{\bm{x}^{(*)}\}, \{1\}, k_{\mathbb{X}})$ with a unique component and uses \ref{eq:inference-probability}, along with  $\rho_{\mathbf x', \mathbf y'}$,  to infer an output KDM $\rho_y$.

\begin{algorithm}[tbh] 
\caption{Discriminative training with bags of instances}
\label{alg:discriminative_train_bags}
\textbf{input:} training dataset $\bm D = \{(\bm{X}^{(i)},\bm{y}^{(i)})\}_{i=1\dots \ell}$ with $\bm X^{(i)}={(\bm{x}^{(i)j}})_{j=1\dots m_i}$,
    kernels  $k_{\mathbb X, \sigma}$ and $k_{\mathbb Y}$
    number of components $m'$, 
    loss function $\mathcal{L}$
\begin{algorithmic}[1]
    \STATE Initialize $\bm{C_{\mathbf {x'y'}}}  \gets  \{(\bm{x'}^{(i)},\bm{y'}^{(i)})\}_{i=1\dots m'}$ with a random sample from $\bm D$
    \STATE \quad $\bm{p}_{\mathbf {x'y'}} \gets (\frac{1}{m'})_{i=1\dots m'}$
    \STATE \quad $\rho_{\mathbf{x',y'}}  \gets  (\bm C_{\mathbf {x'y'}},\bm p_{\mathbf {x'y'}},k_{\mathbb X, \sigma}\otimes k_{\mathbb Y})$
    \STATE \quad $\rho^{(i)}_\mathbf{x} \gets (\bm{X}^{(i)}, (\frac{1}{m_i},\dots,\frac{1}{m_i}), k_{\mathbb{X}})$ for all $i=1\dots \ell$ 
    \REPEAT 
    \STATE Calculate $\rho^{(i)}_\mathbf{y}$ using  Eq. \ref{eq:inference-probability} with parameters $\rho^{(i)}(\mathbf{x})$ and  $\rho({\mathbf{x',y'}})$ for all $i=1\dots \ell$
    \STATE Perform a step of gradient-based method to optimize:
    $                \min_{\bm C_{\mathbf{x',y'}} , \bm p_{\mathbf{x',y'}} ,\sigma} \sum_{i=1}^n  \mathcal{L}(\rho_{\mathbf y}^{(i)}, \bm{y}^{(i)})$
    
    \UNTIL stop criteria is met
    \STATE \textbf{return} $\rho_{\mathbf{x',y'}} = (\bm C_{\mathbf {x'y'}},\bm p_{\mathbf {x'y'}},k_{\mathbb X}\otimes k_{\mathbb Y})$
\end {algorithmic}
\end {algorithm}

We compared this approach with a state-of-the-art LLP method  based on a mutual contamination framework (LMMCM) \citep{Scott2020LearningFramework}. We followed the same experimental setup as the one used by \citep{Scott2020LearningFramework}, where the authors built a set of bags of different sizes with the corresponding label proportions from a conventional classification dataset. The elements of the bags are sampled in such a way that the label proportions are iid uniform on $[0,\frac{1}{2}]$ and on $[\frac{1}{2}, 1]$.  The base datasets  are Adult and MAGIC Gamma Ray Telescope from the UCI repository \footnote{\url{http://archive.ics.uci.edu/ml}}.

\begin{sidewaystable} 
\caption{Performance evaluation for the learning with label proportions tasks. Column headers indicate bag size. Values correspond to AUC plus a t-test $99\%$ confidence interval.}

\centering
\begin{tabular}{llllll}
\hline

Dataset, sampling                          & Method          & 8                 & 32                & 128               & 512               \\
\hline
\multirow{2}{*}{Adult, $[0, \frac{1}{2}]$} %& InvCal          & $0.8720\pm0.0035$ & $0.8672\pm0.0067$ & $0.8537\pm0.0101$ & $0.7256\pm0.0159$ \\
                                           %& alter-$\inf$SVM & $0.8586\pm0.0185$ & $0.7394\pm0.0686$ & $0.7260\pm0.0953$ & $0.6876\pm0.1219$ \\
                                           & LMMCM            & $0.8728\pm0.0034$ & $0.8693\pm0.0084$ & $0.8669\pm0.0073$ & $0.8674\pm0.0072$ \\
                                           & KDM             & $0.8797\pm0.0054$ & $0.8786\pm0.0048$ &  $0.8757\pm0.0056$ &  $0.8697\pm0.0038 $ \\
\hline
\multirow{2}{*}{Adult, $[\frac{1}{2},1]$}  %& InvCal          & $0.8680\pm0.0021$ & $0.8598\pm0.0073$ & $0.8284\pm0.0093$ & $0.7480\pm0.0500$ \\
                                           %& alter-$\inf$SVM & $0.8587\pm0.0097$ & $0.7429\pm0.1473$ & $0.8204\pm0.0318$ & $0.7602\pm0.1215$ \\
                                           & LMMCM           & $0.8584\pm0.0295$ & $0.8644\pm0.00937$ & $0.8601\pm0.00811$ & $0.8500\pm0.0335$ \\
                                           & KDM             & $0.8810\pm0.0083$ & $0.8702\pm0.0121$ &  $0.8666\pm0.0081$ &  $0.8601\pm0.0073$ \\
\hline
\multirow{2}{*}{MAGIC, $[0, \frac{1}{2}]$} %& InvCal          & $0.8918\pm0.0076$ & $0.8574\pm0.0079$ & $0.8295\pm0.0139$ & $0.8133\pm0.0109$ \\
                                           %& alter-$\inf$SVM & $0.8701\pm0.0026$ & $0.7704\pm0.0818$ & $0.7753\pm0.0207$ & $0.6851\pm0.1580$ \\
                                           & LMMCM           & $0.8909\pm0.0013$ & $0.8799\pm0.0203$ & $0.8753\pm0.0283$ & $0.8734\pm0.0165$ \\
                                           & KDM             & $0.8957\pm0.00132$ & $0.8766\pm0.0083$ &  $0.8709\pm0.0063$ &  $0.8578\pm0.0043 $ \\
\hline
\multirow{2}{*}{MAGIC, $[\frac{1}{2},1]$}  %& InvCal          & $0.8936\pm0.0066$ & $0.8612\pm0.0056$ & $0.8180\pm0.0092$ & $0.8215\pm0.0136$ \\
                                           %& alter-$\inf$SVM & $0.8689\pm0.0135$ & $0.8219\pm0.0218$ & $0.8179\pm0.0487$ & $0.7949\pm0.0478$ \\
                                           & LMMCM           & $0.8911\pm0.0149$ & $0.8790\pm0.0164$ & $0.8684\pm0.0083$ & $0.8567\pm0.0165$ \\
                                           & KDM             &  $0.8957\pm0.0093$ & $0.8793\pm0.0121$ &  $0.8709\pm0.0058$ &  $0.8578\pm0.045 $
\end{tabular}
\label{tab:LLP-results}
\end{sidewaystable}

Table \ref{tab:LLP-results} shows the evaluation of the LMMCM and the KDM methods using the area under the ROC curve (AUC) metric. The results are reported for different bag sizes, the two data sets, and the $[0,\frac{1}{2}]$ and  $[\frac{1}{2}, 1]$ sampling schemes. The results show that  KDM  has a competitive performance equal to that of a state-of-the-art method explicitly designed for the LLP task. The good performance of  KDM method is a result of its ability to naturally represent the uncertainty of the training samples.

\section{Conclusions}
The KDM framework presents a simple yet effective mechanism for representing joint probability distributions of both continuous and discrete random variables, providing a versatile tool for various machine learning tasks. The exploratory experiments shed light on the potential of this framework and show how it can be applied to diverse tasks such as conditional generation and weakly supervised learning. Quantum density matrices have long been used as the foundation of quantum mechanics, and their ability to blend linear algebra and probability is very attractive as a tool for machine learning. This work illustrates how such a formalism can be seamlessly and efficiently integrated with probabilistic deep learning models, opening up new avenues of research on the intersection of these fields.

\section*{Statements and Declarations}

The authors declare that they have no known competing interests.

%\section{Limitations}
%While our research highlights the potential of KDM in machine learning, it also reveals certain limitations. First, density estimation methods may struggle with high-dimensional data; it is important to investigate how KDM is affected by increasing data dimensions.  Second, the computational demands of integrating quantum density matrices with deep learning models may pose scalability challenges in large-scale applications. Finally, our study, being exploratory in nature, does not fully evaluate the applicability of KDM outside of conditional generation and weakly supervised learning, so this remains to be further explored.

\begin{appendices}

\section{Proofs}
\setcounter{theorem}{0}
\begin{proposition}{} 
Let $\rho_{\mathbf{x}}=(\bm{C},\bm{p},k_\mathrm{cos})$ be a KDM over  $\mathbb{R}^n$; let $\mathbb{X}=\{\bm{b}^{(1)},\dots,\bm{b}^{(n)} \} \subset \mathbb{R}^n$  be an orthogonal basis of $\mathbb{R}^n$, then $\{f_{\rho}(\bm{b}^{(i)})\}_{i=1,\dots n}$ is a categorical probability distribution for the random variable  $\mathbf{x} \in \mathbb{X}$.
\end{proposition}
\begin{proof}
Let $\bm C = \{\bm x^{(j)}\}_{j=1\dots m}$ , then $f_{\rho_{\mathbf{x}}}(\bm b^{(i)}) =\sum_{j=1}^m p_j \frac{\langle \bm b^{(i)},\bm x^{(j)}\rangle^2}{\langle \bm x^{(j)},\bm x^{(j)}\rangle} \ge 0$.
\begin{align*}
    \sum_{i=1}^n f_{\rho_{\mathbf{x}}}(\bm b^{(i)}) &= \sum_{i=1}^n \sum_{j=1}^m p_j \frac{\langle \bm b^{(i)},\bm x^{(j)}\rangle^2}{\langle \bm x^{(j)},\bm x^{(j)}\rangle} \\
    &= \sum_{j=1}^m p_j \sum_{i=1}^n \frac{\langle \bm b^{(i)},\bm x^{(j)}\rangle^2}{\langle \bm x^{(j)},\bm x^{(j)}\rangle} \\
    &= \sum_{j=1}^m p_j \|  \frac{\bm x^{(j)}}{\langle \bm x^{(j)},\bm x^{(j)}\rangle}\|^2 \; (\mathbb X \text { is a basis of } \mathbb R ^n)\\
    &= \sum_{j=1}^m p_j  \\
    &= 1
\end{align*}
\end{proof}

\setcounter{theorem}{2}
\begin{proposition}
Let  $\mathbb{X}=\{\bm{e}^{(1)},\dots,\bm{e}^{(n)} \}$  be the canonical basis of  $\mathbb R^n$; let  $\bm{D}=\{\bm x^{(1)}, \dots,\bm x^{(\ell)}\}\subseteq \mathbb R^n $  be a set of iid samples drawn from $\mathbb X$ with categorical probability distribution $\bm{p}=(p_1,\dots,p_n)$; let   $\rho^\ell_\mathbf{x}=(\bm{D},\bm{p}=(\frac{1}{\ell},\dots,\frac{1}{\ell}),k_{\mathrm{cos}})$  be a KDM; and let $\hat{f}_{\rho^\ell_\mathbf{x}}(\bm x)$ be defined as in  Eq. \ref{eq:f_hat-cos}.  Then $\hat{f}_{\rho_\ell}(\bm e^{(i)}) \rightarrow p_i$ as $\ell \rightarrow \infty$.
\end{proposition}

\begin{proof}
\begin{align*}
     f_{\rho_{\mathbf{x}}}(\bm e^{(i)}) &=  \sum_{j=1}^\ell  \frac{1}{\ell} k_{\text{cos}}^2(\bm x^{(j)},\bm e^{(i)}) \\
     &=  \sum_{j=1}^\ell  \frac{1}{\ell} \langle\bm x^{(j)},\bm e^{(i)}\rangle^2 \\
     &=  \sum_{j=1}^\ell \frac{1}{\ell} 1(\bm x^{(j)} = \bm e^{(i)}) \\
\end{align*}
Because of the large numbers law:
\begin{equation}
\begin{split}
    \lim_{\ell \rightarrow \infty} f_{\rho_{\mathbf{x}}}(\bm e^{(i)}) &= \lim_{\ell \rightarrow \infty} \sum_{j=1}^\ell \frac{1}{\ell} 1(\bm x^{(j)} = \bm e^{(i)}) \\ 
    & = E[1(\bm x^{(j)} = \bm e^{(i)})] \\
    & = p_i
\end{split}
\end{equation}

\end{proof}

\begin{proposition}
    Let $\mathbf{x}'$ and $\mathbf{y}'$ be random variables with a joint probability distribution represented by a KDM   $\rho_{\mathbf{x',y'}}$ (Eq. \ref{eq:rho_xy}), and let  $\rho_{\mathbf{x}}$ be the KDM defined by  Eq. \ref{eq:rho_x}), then the  KDM $\rho_{\mathbf{y}}$ (Eq. \ref{eq:rho_y} and Eq. \ref{eq:inference-probability}) represents a predicted probability distribution with PDF:
    \begin{equation}
    \hat{f}_{\rho_{\mathbf{y}}}(\bm{y}) = \sum_{\ell=1}^m p_\ell \hat{f}_{\rho_{\mathbf{x',y'}}}(\bm y|\bm x^{(\ell)}) \notag
\end{equation}
    where 
    \begin{itemize}
        \item $\hat{f}_{\rho_{\mathbf{x',y'}}}(\bm y|\bm x^{(\ell)})=\frac{\hat{f}_{\rho_{\mathbf{x',y'}}}(\bm x^{(\ell)}, \bm y)}{\int \hat{f}_{\rho_{\mathbf{x',y'}}}(\bm x^{(\ell)}, \bm y)d\bm y}$   for  $k_\mathbb Y = k_\text{rbf}$
\item $\hat{f}_{\rho_{\mathbf{x',y'}}}(\bm y|\bm x^{(\ell)})=\frac{\hat{f}_{\rho_{\mathbf{x',y'}}}(\bm x^{(\ell)}, \bm y)}{\sum_{i=1}^n \hat{f}_{\rho_{\mathbf{x',y'}}}(\bm x^{(\ell)}, \bm e^{(i)})}$   for  $k_\mathbb Y = k_\text{cos}$
    \end{itemize}

\end{proposition}
\begin{proof}

\textbf{Step 1}. Show that for both   $k_\mathbb Y = k_\text{rbf}$ and  $k_\mathbb Y = k_\text{cos}$ 
\begin{equation}\label{eq:proof-hat_f-cond}
    \hat{f}_{\rho_{\mathbf{x',y'}}}(\bm y|\bm x^{(\ell)})=\frac{\hat{f}_{\rho_{\mathbf{x',y'}}}(\bm x^{(\ell)}, \bm y)}{\mathcal{M}_{k_{\mathbb X}}\sum_{j=1}^{m'} p'_j  k^2_\mathbb{X}(\bm x^{(\ell)}, \bm x'^{(j)})}
\end{equation}

The case  $k_\mathbb Y = k_\text{rbf}$  follows from Eq. \ref{eq:marginal-kernel-mixture}. For the case $k_\mathbb Y = k_\text{cos}$  we have
\begin{align*}
   & \sum_{i=1}^n \hat{f}_{\rho_{\mathbf{x',y'}}}(\bm x^{(\ell)}, \bm e^{(i)})  \\ 
    = & \sum_{i=1}^n \sum_{j=1}^{m'} p'_j \mathcal{M}_{k_{\mathbb X}} \mathcal{M}_{k_\text{cos}} k^2_\mathbb{X}(\bm x^{(\ell)},\bm x'^{(j)})k^2_\text{cos}( \bm e^{(i)},\bm y'^{(j)}) \\
    = & \mathcal{M}_{k_{\mathbb X}} \sum_{j=1}^{m'} p'_j   k^2_\mathbb{X}(\bm x^{(\ell)},\bm x'^{(j)}) \sum_{i=1}^n  k^2_\text{cos}( \bm e^{(i)},\bm y'^{(j)}) \\
    = & \mathcal{M}_{k_{\mathbb X}} \sum_{j=1}^{m'} p'_j   k^2_\mathbb{X}(\bm x^{(\ell)},\bm x'^{(j)})
\end{align*}
\textbf{Step 2}, Use the derivation above to prove the main result:
\begin{align*}
   & \hat{f}_{\rho_{\mathbf{y}}}(\bm{y}) \\ 
   = & \mathcal{M}_{k_{\mathbb Y}} \sum_{i=1}^{m'} p''_i   k^2_\mathbb{Y}(\bm y,\bm y'^{(i)}) &\text{(use Eq. \ref{eq:f_hat-density})}\\
    = & \mathcal{M}_{k_{\mathbb Y}} \sum_{i=1}^{m'} \sum_{\ell=1}^m\frac{p_\ell p'_i 
    k^2_{\mathbb X}(\bm{x}^{(\ell)},\bm{x}'^{(i)})}{\sum_{j=1}^{m'}  p'_j k^2_{\mathbb X}(\bm{x}^{(\ell)},\bm{x}'^{(j)})}   k^2_\mathbb{Y}(\bm y,\bm y'^{(i)}) &\text{(use Eq. \ref{eq:inference-probability})}\\
    = &  \sum_{\ell=1}^m p_\ell \frac{ \mathcal{M}_{k_{\mathbb X}} \mathcal{M}_{k_{\mathbb Y}} \sum_{i=1}^{m'} p'_i k^2_{\mathbb X}(\bm{x}^{(\ell)},\bm{x}'^{(i)})k^2_\mathbb{Y}(\bm y,\bm y'^{(i)})}{\mathcal{M}_{k_{\mathbb X}}\sum_{j=1}^{m'}  p'_j k^2_{\mathbb X}(\bm{x}^{(\ell)},\bm{x}'^{(j)})}    &\\
    = & \sum_{\ell=1}^m p_\ell \frac{\hat{f}_{\rho_{\mathbf{x',y'}}}(\bm x^{(\ell)}, \bm y)}{\mathcal{M}_{k_{\mathbb X}}\sum_{j=1}^{m'} p'_j  k^2_\mathbb{X}(\bm x^{(\ell)}, \bm x'^{(j)})} &\\
    = & \sum_{\ell=1}^m p_\ell \hat{f}_{\rho_{\mathbf{x',y'}}}(\bm y|\bm x^{(\ell)}) &\text{(use Eq. \ref{eq:proof-hat_f-cond})}
\end{align*}
\end{proof}

\section{Experimental Setup}
In this apendix, we provide a comprehensive description of our experimental setup to ensure reproducibility.
\subsection{Hardware Specifications}
 Our experiments were conducted using a server that incorporates a 64-core Intel Xeon Silver 4216 CPU 2.10 GHz processor, 128 GB of RAM, and two NVIDIA RTX A5000 GPUs.

\subsection{Classification with kernel density matrices}
Three different models were assessed in this experiment as explained in Subsection \ref{subsec:classification_with_quantum_kernel_mixtures}: KDM classification model (KDM), KDM model fined-tuned for a generation with maximum likelihood learning (ML-KDM), and a baseline model with the same encoder coupled with a dense layer. 
The encoder model utilized for both MNIST and Fashion-MNIST datasets was identical. It consisted of the following components:
\begin{itemize}
    \item Lambda Layer: The initial layer converted each sample into a 32-float number and subtracted 0.5 from it.
    \item Convolutional Layers: Two convolutional layers were appended, each with 32 filters, 5-kernel size, same padding, and strides 1 and 2, respectively. Subsequently, two more convolutional layers were added, each with 64 filters, 5-kernel size, same padding, and strides 1 and 2, respectively. Finally, a convolutional layer with 128 filters, 7-kernel size, and stride 1 was included. All convolutional layers employed the Gaussian Error Linear Unit (GELU) activation function.
    \item Flattened Layer: The output from the previous layer was flattened.
    \item Dropout Layer: A dropout layer with a dropout rate of 0.2 was introduced.
    \item Dense Layer: The subsequent dense layer's neuron encoding size was determined using a grid hyperparameter search in the range of $2^i$, where $i \in \{1,\cdots,7\}$.
\end{itemize}

The decoder architecture consisted of the following components:

\begin{itemize}
    \item Reshape Layer: The input to the decoder passed through a reshape layer.
    \item Convolutional Layers: Three convolutional layers were employed, each with 64 filters and kernel sizes of seven, five, and five, respectively. The strides for these layers were set to 1, 1, and 2, while the padding was configured as valid, same, and same, respectively.
    \item Additional Convolutional Layers: Subsequently, three more convolutional layers were utilized, each with 32 filters, 5-kernel size, and strides 1, 2, and 1, respectively. All these layers utilized the Gelu activation function and had padding set to "same."
    \item Final Convolutional Layer: The decoder concluded with a 1-filter convolutional layer employing same padding.
\end{itemize}

Similar to the encoder, all layers in the decoder employed the GELU activation function. 

For Cifar-10 a different encoder was used for 
\begin{itemize}
    \item Convolutional Layers: The input tensor is passed through a series of convolutional layers. Each layer applies a 3x3 kernel to extract features from the input. The activation function used is the GELU. Padding is set to 'same'. Batch normalization is applied after each convolutional layer to normalize the activations.
    \item Max Pooling: After each pair of convolutional layers, max pooling is performed using a 2x2 pool size.
    \item Encoding Layer: The feature maps obtained from the previous layers are flattened into a vector representation using the Flatten layer.
    \item Dropout: A dropout layer with a dropout rate of 0.2 is added.
    \item Hidden Layer: A dense layer of variable size is introduced to further transform the encoded features. The size is search  
    \item Dropout: Another dropout layer with a dropout rate of 0.2 is added after the hidden layer
\end{itemize}

The decoder architecture for Cifar-10 consisted of the following components:
\begin{itemize}
    \item Dense Layer: The input tensor is passed through a dense layer that restores the tensor to its original size.
    \item Reshape Layer: The tensor is reshaped to match the spatial dimensions of the original input using the Reshape layer.
    \item Convolutional Transpose Layers: The reshaped tensor undergoes a series of convolutional transpose layers. Each layer applies a 3x3 kernel to upsample the feature maps. The activation function used is 'gelu'. Padding is set to 'same' to maintain the spatial dimensions.
    \item Batch Normalization: Batch normalization is applied after each convolutional transpose layer to normalize the activations.
    \item Upsampling: UpSampling2D layers with a size of (2, 2) are used to increase the spatial dimensions of the feature maps.
    \item Final Convolutional Transpose Layer: The last convolutional transpose layer reconstructs the original number of channels in the input data. Padding is set to 'same'.
    \item Activation Function: The final output is passed through an activation function, \'sigmoid\', to ensure the reconstructed data is within the appropriate range.
\end{itemize}

The number of components in every dataset for the KDM classification model was searched in $2^i$, where $i \in \{2,\cdots,10\}$. The same for ML-KQM. For the baseline model, the dense layer was searched using the same number of parameters that the KDM classification model generates. The best parameters can be found in Table \ref{table:best_parameters_experiment_classification}.

\begin{sidewaystable*}[htbp]
\caption{Best parameters for the classification with kernel density matrix experiment.} \label{table:best_parameters_experiment_classification}
\begin{tabular}{lllllll}
\hline
\hline
Dataset      & \multicolumn{2}{l}{KDM} & \multicolumn{2}{l}{KDM-ML} & \multicolumn{2}{l}{Baseline Dense Layer} \\
\hline
             & Number of    & Encoded  & Number of     & Encoded    & Number of            & Encoded           \\
             & Components   & Size     & Components    & Size       & Components           & Size              \\
             
\hline
Mnist        & 512          & 128      & 512           & 128        & 8                    & 8                 \\
Fashin-Mnist & 256          & 128      & 256           & 128        & 4                    & 16                \\
Cifar-10     & 1024         & 32       & 1024          & 32         & 4                    & 64           \\    
\hline
\end{tabular}
\end{sidewaystable*}

The results reported in Table \ref{tab:classification-results} were generated by running each algorithm in each dataset ten times and averaging their results.  

\subsection{Classification with label proportions}
The experimental setup of this experiment is based on the work of  Scott et al. \cite{Scott2020LearningFramework}. The study utilizes two datasets, namely Adult (T = 8192) and MAGIC Gamma Ray Telescope (T = 6144), both obtained from the UCI repository. \footnote{http://archive.ics.uci.edu/ml.}. For each dataset, the authors created datasets with bags of instances with different proportions of LPs.  Each dataset has its distribution of label proportions (LPs), and a bag size parameter, denoted as 'n'. The total number of training instances denoted as 'T', is fixed for each dataset, resulting in T/n bags. The LPs for the experiments are independently and uniformly distributed within the ranges [0, 1/2] and [1/2, 1]. The bag sizes considered are n = \{8, 32, 128, and 512\}. Consequently, there are a total of 16 experimental configurations (2 LP distributions x 2 datasets x 4 bag sizes). For each configuration, 5 different pairs of training and test sets were created. Results are reported as the average of the  AUC over the five test sets of each configuration. The numerical features in both datasets are standardized to have a mean of 0 and a variance of 1, while the categorical features are encoded using one-hot encoding. Hyperparameters are found using a validation subset (10\%) of each training set. The best hyperparameters for each configuration are reported in Table \ref{tab:hyperparameters-llp}

\begin{table}[htbp]
  \caption{Best hyper-parameters for the learning with label proportion experiment using KDM Model.}
  \label{tab:hyperparameters-llp}
  \centering
  \begin{tabular}{ccccc}
    \toprule
    Dataset, sample & Bag Size & Num. Components & Learning Rate \\
    \midrule
    Adult, $[0, \frac{1}{2}]$ & 8 & 32 & 0.005 \\
    Adult, $[0, \frac{1}{2}]$ & 32 & 16 & 0.001 \\
    Adult, $[0, \frac{1}{2}]$ & 128 & 512 & 0.001 \\
    Adult, $[0, \frac{1}{2}]$ & 512 & 64 & 0.005 \\
    Adult, $[\frac{1}{2},1]$ & 8 & 16 & 0.005 \\
    Adult, $[\frac{1}{2},1]$ & 32 & 64 & 0.001 \\
    Adult, $[\frac{1}{2},1]$ & 128 & 128 & 0.001 \\
    Adult, $[\frac{1}{2},1]$ & 512 & 64 & 0.005 \\
    MAGIC, $[0, \frac{1}{2}]$ & 8 & 256 & 0.005 \\
    MAGIC, $[0, \frac{1}{2}]$ & 32 & 128 & 0.005 \\
    MAGIC, $[0, \frac{1}{2}]$ & 128 & 128 & 0.001 \\
    MAGIC, $[0, \frac{1}{2}]$ & 512 & 256 & 0.001 \\
    MAGIC, $[\frac{1}{2},1]$ & 8 & 16 & 0.005 \\
    MAGIC, $[\frac{1}{2},1]$ & 32 & 128 & 0.005 \\
    MAGIC, $[\frac{1}{2},1]$ & 128 & 32 & 0.005 \\
    MAGIC, $[\frac{1}{2},1]$ & 512 & 128 & 0.001 \\
    \bottomrule
  \end{tabular}
\end{table}

\subsection{Code}

The code repository at \url{https://github.com/fagonzalezo/kdm_for_probabilistic_DL_experiments} provides detailed instructions to reproduce the results presented in this manuscript. 

For conducting experiments involving learning with label proportions, as mentioned above, we have relied on the work of Scott et al. \cite{Scott2020LearningFramework}. The authors made the code available at this GitHub repository \url{https://github.com/Z-Jianxin/Learning-from-Label-Proportions-A-Mutual-Contamination-Framework}. 




\end{appendices}

\bibliography{references}


%%\bibliography{sn-bibliography}% common bib file

\end{document}